\documentclass[conference]{IEEEtran}
\IEEEoverridecommandlockouts

\usepackage{cite}
\usepackage{amsmath,amssymb,amsfonts}
\usepackage{algorithm}
\usepackage{algpseudocode}
\usepackage{graphicx}
\usepackage{textcomp}
\usepackage{xcolor}
\usepackage{booktabs}
\usepackage{url}
\usepackage{diagbox}
\usepackage{fancyvrb}
\usepackage{multirow}
\usepackage{tikz}
\usepackage{orcidlink}
\usetikzlibrary{positioning,arrows.meta,shapes.geometric,fit,backgrounds,calc}

\def\BibTeX{{\rm B\kern-.05em{\sc i\kern-.025em b}\kern-.08em
    T\kern-.1667em\lower.7ex\hbox{E}\kern-.125emX}}

\begin{document}

\title{Multilingual Fact-Checking at Scale: Fine-Tuned Compact Models vs LLMs}

\author{\IEEEauthorblockN{Pratuat Amatya\,\orcidlink{0009-0004-2593-056X}}
\IEEEauthorblockA{\textit{Factiverse AS and University of Stavanger}\\
Stavanger, Norway \\
pratuat@factiverse.ai}
\and
\IEEEauthorblockN{Vinay Setty\,\orcidlink{0000-0002-9777-6758}}
\IEEEauthorblockA{\textit{Factiverse AS and University of Stavanger}\\
Stavanger, Norway \\
vinay@factiverse.ai}
}

\maketitle

\begin{abstract}
We present a multilingual fact-checking system deployed at Factiverse, designed for high-throughput and low-latency operation across diverse languages. The system follows a modular pipeline with three stages: claim detection, evidence retrieval and re-ranking, and veracity prediction. We fine-tune XLM-RoBERTa-Large for claim detection, mmBERT-base for three-label stance classification (Supports/Refutes/Mixed), and a SetFit-based multilingual re-ranker for claim--evidence matching. We compare these components against strong LLM baselines, including GPT-5.2, Claude Opus~4.6, and Qwen3-8b. Experiments on production data spanning 114 languages for claim detection and 28 languages for veracity prediction show that task-specific fine-tuning provides strong and stable multilingual performance, while the fine-tuned retrieval model remains competitive with modern proprietary embeddings. Same-hardware latency measurements further show large efficiency gains for encoder-based components, supporting their use in production deployments with tight cost and privacy constraints. Overall, compact fine-tuned, self-hosted models remain a practical and effective foundation for multilingual fact-checking at scale. Code and data used for this study are available at \url{https://github.com/factiverse/factcheck-editor}.
\end{abstract}

\begin{IEEEkeywords}
multilingual fact-checking, claim detection, natural language inference, evidence retrieval, fine-tuning
\end{IEEEkeywords}

\section{Introduction}

Online misinformation continues to scale across platforms, languages and domains, creating sustained pressure on professional fact-checkers and newsrooms. Recent measurements indicate that false or misleading narratives now propagate orders of magnitude faster than fact-checks themselves, and across far more languages than any single organisation can manually cover~\cite{Botnevik:2020:SIGIRa,setty2024surprising}. Automated support systems are therefore no longer optional but operationally necessary. However, building reliable industrial-grade fact-checking systems remains challenging for at least four reasons: (i)~the long tail of languages and orthographies makes monolingual or even bilingual approaches inadequate; (ii)~claims often require numerical, temporal or multi-hop reasoning over heterogeneous web evidence; (iii)~newsrooms operate under strict latency, cost and privacy constraints that rule out per-query reliance on hosted closed-source APIs; and (iv)~the training signal is fundamentally imbalanced, since most sentences in any document are not check-worthy and most retrieved passages are not relevant.

Recent progress in large language models (LLMs) has renewed optimism about end-to-end fact-checking~\cite{vykopal2024generative}. Models such as GPT-5.x, Claude Opus~4.x, Qwen3.x and Gemma4 exhibit strong general reasoning capabilities and broad multilingual coverage~\cite{gao2023rarr}, and an emerging line of work argues that retrieval-augmented LLMs alone could subsume the classical claim$\rightarrow$evidence$\rightarrow$verdict pipeline. Nevertheless, practical deployment requires consistent performance across languages, stable precision--recall trade-offs, and self-hostable solutions compatible with newsroom and fact-checker privacy requirements. This is especially important in agentic fact-checking workflows where a large LLM acts as an orchestrator and calls specialist tools: compact encoder models are attractive as callable agents for high-throughput classification and retrieval because they are fast, deterministic, and inexpensive to run at scale. In this paper we ask, concretely, whether \emph{compact fine-tuned encoder models still beat prompt-based LLM pipelines for industrial multilingual fact-checking}, and we answer in the affirmative across the three core components.

In this work, we analyse three core components of an industrial fact-checking pipeline: claim detection, retrieval re-ranking, and veracity prediction~\cite{Guo:2022:JACL,setty2024surprising}, deployed as a modular system at Factiverse and evaluated against the strongest current proprietary LLMs. For claim detection and stance, we evaluate multilingual Transformer models across a broad set of languages. A fine-tuned XLM-RoBERTa-Large model achieves consistently strong macro F1 scores, outperforming large proprietary LLMs particularly in low-resource and morphologically rich languages. These results highlight the effectiveness of task-specific fine-tuning for multilingual claim identification. For retrieval re-ranking, we compare embedding-based bi-encoders fine-tuned vs untuned embedding models both open source and proprietary models. Overall, the results show that carefully fine-tuned Transformer architectures remain highly competitive, and in several practical settings preferable, to large general-purpose LLMs for production-grade fact-checking systems, including agentic settings where an LLM controller delegates repeated verdict and retrieval decisions to specialist encoder tools.

An additional constraint in real-world deployments concerns data governance. Journalists and media organisations often cannot transmit sensitive material to external APIs. The evaluated cross-encoder architecture can be fully self-hosted, enabling privacy-preserving deployment while maintaining high effectiveness.

\paragraph{Contributions.} Concretely, this paper makes the following contributions:
\begin{itemize}
    \item A description and ablation of the deployed Factiverse multilingual fact-checking pipeline, covering claim detection, evidence retrieval with question decomposition, cross-lingual re-ranking via SetFit, and veracity prediction via stance aggregation (Section~\ref{sec:method}).
    \item A systematic comparison of fine-tuned models (XLM-RoBERTa-Large for claim detection and mmBERT-base for veracity prediction) against three state-of-the-art LLMs (GPT-5.2, Claude Opus~4.6, Qwen3-8b) across 114 languages for claim detection and 28 languages for veracity prediction, using identical prompts and held-out human-annotated data (Section~\ref{sec:eval}).
    \item Concrete fine-tuning recipes (loss formulation, class re-balancing, multilingual data augmentation, verdict-phrase scrubbing) that explain \emph{why} compact encoder models retain a measurable advantage over zero/few-shot LLMs on production data, supported by per-language and per-class error analysis.
    \item Practical evidence that a 560M-parameter encoder, fully self-hostable, is competitive with multi-hundred-billion-parameter API models while satisfying newsroom-grade latency, cost, and privacy constraints.
\end{itemize}

The rest of the paper is organised as follows. Section~\ref{sec:related} reviews related work on automated fact-checking. Section~\ref{sec:method} describes the Factiverse pipeline and the fine-tuning methodology for each component. Section~\ref{sec:eval} presents the experimental setup, baselines, and prompts. Section~\ref{sec:results} reports cross-lingual results and a detailed error analysis, before concluding in Section~\ref{sec:conclusion}.

\section{Related Work}
\label{sec:related}

Research on automated fact-checking spans three main components: claim detection, evidence retrieval, and veracity prediction. We discuss each in turn, then position our contribution against the recent surge of LLM-based end-to-end approaches.

\paragraph{Claim detection.} Claim detection has primarily been modelled as supervised sentence-level classification in political and news domains, beginning with ClaimBuster~\cite{Hassan:2017:VLDB} and continuing through the CLEF CheckThat! shared tasks~\cite{Alam:2021:arXiv}. Recent multilingual surveys~\cite{Panchendrarajan:2024:arXiv} highlight a persistent gap: most labelled data is English-language, and zero-shot transfer to morphologically richer languages is brittle for models that lack genuinely multilingual pretraining. A complementary line studies what makes a claim ``check-worthy'' beyond mere factuality (relevance, harm, salience), but the evaluation infrastructure remains dominated by binary classification benchmarks.

\paragraph{Evidence retrieval.} Evidence retrieval is typically framed as an open-domain retrieval task over large and heterogeneous corpora, with hybrid lexical-plus-dense pipelines now standard in industrial systems~\cite{10.1145/3701716.3715300}. Modern bi-encoders such as those from OpenAI, Google, Perplexity and Qwen offer strong English performance, but cross-lingual evaluations remain rare. Recent work on contrastive sentence embeddings~\cite{Reimers:2019:Arxiv,setfit} demonstrates that small amounts of in-domain relevance feedback can sharply improve performance over general-purpose embeddings; we build directly on this insight.

\paragraph{Veracity prediction.} Veracity prediction, often cast as natural language inference or stance classification, determines whether evidence supports or refutes a claim~\cite{Popat::a,schlichtkrull2023averitec,Guo:2022:JACL}. Benchmarks such as FEVER, MultiFC and AVeriTeC have driven steady progress, but most operate on relatively clean evidence and underrepresent the noisy, multilingual web data encountered in production. Real-world fact-check explanations also typically contain explicit verdict phrases that, if not removed, leak label information into the model; we address this preprocessing concern explicitly in Section~\ref{sec:method}.

Graph-based formulations have also proven effective, including stream-based reasoning over knowledge graphs for fact checking and hierarchical graph inference methods for evidence-aware verification~\cite{Shiralkar:2017:ICDM,Mao:2022:ICDM}; related work on rumor detection in heterogeneous graphs further motivates graph-structured modeling choices in misinformation settings~\cite{Yuan:2019:ICDM}.

\paragraph{End-to-end LLM fact-checking.} The rise of large language models has led to end-to-end approaches that combine retrieval, reasoning, and explanation generation within a single generative framework. Search-augmented generation and self-verification methods aim to reduce hallucinations and improve factual grounding~\cite{gao2023rarr,Manakul:2023:arXiv,Chern:2023:arXiva,Mishra:2024:arXiva,Wang:2023:arXiv}. However, deployment in practice remains limited due to latency, cost, hallucination risks, and data governance constraints, particularly in newsroom and enterprise settings~\cite{oshikawa2020survey,quelle2024perils}. Several papers report that while LLMs produce fluent justifications, their per-claim verdicts are inconsistent across reruns and degrade sharply in low-resource languages; we corroborate this observation quantitatively in Section~\ref{sec:results}.

\paragraph{Industrial pipelines.} In contrast, industrial systems commonly adopt modular pipelines that separate retrieval, ranking, and classification~\cite{Botnevik:2020:SIGIRa,venktesh2025livefc,vatndal2025shortcheck}. Such designs allow better control of precision--recall trade-offs and easier integration with curated sources. Factiverse has previously introduced commercial fact-checking tools and multilingual benchmarks~\cite{setty2024factcheck,setty2024surprising}. Extending this line of work, we compare modern proprietary and open embedding models with a domain-adapted fine-tuned embedding model in multilingual, production-oriented settings, and we provide a per-language error analysis that quantifies \emph{when} a fine-tuned compact encoder remains preferable to large generalist LLMs.

\section{Fact-checking pipeline}
\label{sec:method}

\begin{figure}[t]
\centering
\resizebox{\columnwidth}{!}{%
\begin{tikzpicture}[
  font=\scriptsize,
  node distance=4mm and 5mm,
  every node/.style={align=center},
  box/.style={
    rectangle, rounded corners=2pt, draw=black!70, thick,
    minimum height=8.5mm, minimum width=16mm, inner sep=1.5pt,
    fill=white,
  },
  ft/.style={box, fill=blue!8, draw=blue!60},
  llm/.style={box, fill=orange!15, draw=orange!70},
  io/.style={
    rectangle, draw=black!70, thick,
    minimum height=8.5mm, minimum width=14mm,
    fill=gray!10,
  },
  stage/.style={
    rectangle, rounded corners=4pt,
    draw=black!40, dashed, thick,
    fill=black!2,
    inner xsep=4mm, inner ysep=3mm,
  },
  arr/.style={-{Latex[length=1.8mm]}, thick, black!70},
]

\node[io] (input) {Input\\text};

\node[box, below=8mm of input] (seg) {Sentence\\seg.+coref};
\node[llm, below=4mm of seg] (segllm) {Gemma4\\E4b-it};
\node[ft, right=of seg] (cd) {Claim det.\\\textbf{XLM-R}};
\node[io, right=of cd] (claim) {Check-worthy\\claim};

\node[llm, below=24mm of seg] (qgen) {Question\\gen.};
\node[box, right=of qgen] (search) {Web\\search};
\node[box, below=4mm of search, text width=22mm] (sources)
  {Google $\bullet$ DDG\\Wikipedia $\bullet$ Semantic Scholar\\FactiSearch};
\node[ft, right=of search] (rerank) {Re-ranker\\\textbf{SetFit}};
\node[io, right=of rerank] (passages) {Top-$k$\\passages};

\node[llm, below=35mm of qgen] (summary) {Summary\\+corr.};
\node[ft, right=of summary] (stance) {Stance\\\textbf{mmBERT}};
\node[box, right=of stance] (aggr) {Agg.};
\node[io, right=of aggr] (verdict) {Verdict\\(S/R/M)};

\begin{scope}[on background layer]
\node[stage, fit=(seg)(segllm)(cd)(claim),
  label={[anchor=south,font=\scriptsize\bfseries,yshift=0mm]north:Stage 1: Claim Detection}] {};
\node[stage, fit=(qgen)(sources)(rerank)(passages),
  label={[anchor=south,font=\scriptsize\bfseries,yshift=0mm]north:Stage 2: Retrieval \& Re-ranking}] {};
\node[stage, fit=(summary)(stance)(aggr)(verdict),
  label={[anchor=south,font=\scriptsize\bfseries,yshift=0mm]north:Stage 3: Veracity}] {};
\end{scope}

\draw[arr] (input.south) -- (seg.north);
\draw[arr] (segllm.north) -- (seg.south);
\draw[arr] (seg) -- (cd);
\draw[arr] (cd) -- (claim);

\draw[arr] (claim.south) -- (qgen.north);
\draw[arr] (qgen) -- (search);
\draw[arr] (sources.north) -- (search.south);
\draw[arr] (search) -- (rerank);
\draw[arr] (rerank) -- (passages);

\draw[arr] (passages.south) -- (summary.north);
\draw[arr] (summary) -- (stance);
\draw[arr] (stance) -- (aggr);
\draw[arr] (aggr) -- (verdict);

\end{tikzpicture}
}
\caption{The deployed Factiverse fact-checking pipeline. Blue nodes are fine-tuned encoder components: XLM-RoBERTa-Large for claim detection, SetFit XLM-R for re-ranker, and mmBERT-base for stance classifier; orange nodes are off-the-shelf LLMs used for utility tasks (sentence/coreference cleanup, question decomposition, justification summarisation). Grey nodes denote intermediate artefacts. The verdict is produced by aggregating per-passage stance decisions, and an LLM-generated justification accompanies each prediction.}
\label{fig:pipeline}
\end{figure}
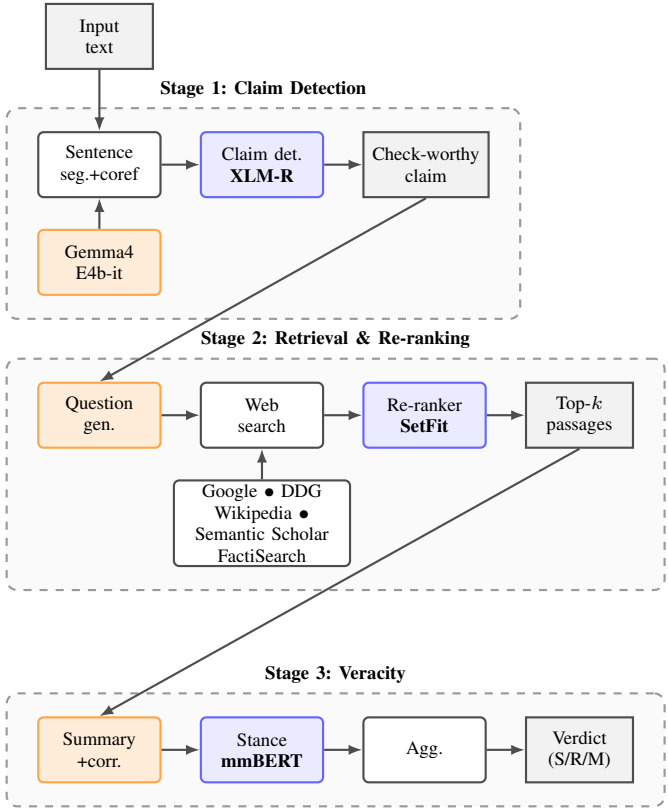

We follow a three-stage fact-checking pipeline that is widely used in the literature~\cite{Guo:2022:JACL} and industry~\cite{setty2024surprising,venktesh2025livefc,vatndal2025shortcheck,Botnevik:2020:SIGIRa}, and instantiate it as the deployed Factiverse system shown in Figure~\ref{fig:pipeline}. It first detects check-worthy statements, referred to as claims, from user-written or LLM-generated text. For each claim, relevant evidence is retrieved from the open web and previously verified fact-checks, and the claim is then verified against this evidence. The pipeline is implemented as a Python FastAPI service deployed on a Kubernetes cluster with auto-scaling on Azure cloud platform, and integrates dedicated models for claim detection, evidence retrieval and ranking, and veracity prediction.

At a high level, the pipeline takes a raw document (e.g, news articles, video/podcast transcripts, LLM generated content) as input (Figure~\ref{fig:pipeline}, left) and emits, for every check-worthy sentence, a three-way veracity verdict (\textsc{Supports}, \textsc{Refutes}, \textsc{Mixed}) together with a natural-language justification (right). Three task-specific encoder models are deployed: XLM-RoBERTa-Large for claim detection and retrieval re-ranking, and mmBERT-base for stance classification, while utility steps such as coreference resolution, question decomposition and the final summary generation are delegated to off-the-shelf LLMs. This split is deliberate: the encoder components are the ones whose accuracy and latency need to be guaranteed across 100+ languages, so they are fine-tuned on production data; the LLM components are used only where their generative flexibility is genuinely required, and can be swapped without retraining the rest of the pipeline.

The remainder of this section describes each stage in detail, focusing on the fine-tuning methodology that yields the performance reported in Section~\ref{sec:results}.

\subsection{Claim Detection}
\label{sec:claim_detection}
We first segment the sentences in a given text and use an open LLM (Gemma4-E4b-it)\footnote{Self-hosted with Ollama} to resolve and decontextualize them (for example, coreference resolution). We next classify these sentences into check-worthy and not-check-worthy. We follow the definition of check-worthy claims from \cite{Panchendrarajan:2024:arXiv}.

We fine-tune the \textit{XLM-RoBERTa-Large} model \footnote{\url{https://huggingface.co/FacebookAI/xlm-roberta-large}}~\cite{DBLP:journals/corr/abs-1911-02116}, using the data from ClaimBuster~\cite{Hassan:2017:VLDB} and CLEF CheckThat Lab!~\cite{Alam:2021:arXiv} along with a dataset collected from Factiverse production system (see \cite{setty2024surprising} for more details) to classify sentences into `Check-worthy' and `Not-check-worthy'.

The fine-tuning objective is sequence-level binary cross-entropy on the \texttt{[CLS]} token, with a linear classification head added on top of the pre-trained encoder. Because the training data is heavily skewed toward non-check-worthy sentences, we apply class-weighted loss with a positive-class weight of $5{:}1$ to reduce false negatives that would otherwise propagate through the rest of the pipeline. The multilingual representations of XLM-RoBERTa-Large allow the resulting classifier to transfer zero-shot to 100+ additional languages without explicit training data, which we evaluate in Section~\ref{sec:eval}.

We use XLM-RoBERTa-Large for claim detection because this stage operates on sentence-level inputs where a 512-token window is sufficient in practice, while still benefiting from XLM-R's strong multilingual representations.

\subsection{Evidence Retrieval}

The second stage retrieves evidence relevant to each detected claim. Since issuing the claim verbatim as a query often yields limited or overly specific results, we first decompose it into targeted verification questions generated by an LLM (Gemma4-E4b-it). These questions are used to query multiple sources, including Google, DuckDuckGo, Wikipedia, Semantic Scholar, and Factiverse's continuously updated fact-check archive\footnote{https://factisearch.ai}. Retrieved documents are deduplicated through approximate matching over URLs, titles, and content. Publicly accessible articles are then scraped and segmented into passages. Both the passages and the queries are encoded using a multilingual sentence encoder~\cite{Reimers:2019:Arxiv}, and passages exceeding a cosine similarity threshold of \textbf{0.6} are retained as candidate evidence. We fine-tune an embedding model using a SetFit classifier \cite{setfit} on relevance feedback collected internally at Factiverse (claim, evidence) pairs.

Fine-tuning follows the two-stage SetFit recipe~\cite{setfit}: in the first stage, the sentence encoder is adapted with a contrastive cosine-similarity loss over RELEVANT/NOT\_RELEVANT pairs, where positives are drawn from human-confirmed relevance feedback and hard negatives are mined from passages that share lexical overlap with the claim but were judged irrelevant. In the second stage, a lightweight logistic classification head is trained on the resulting embeddings to produce a calibrated relevance score. Because human-labelled pairs are scarce in low-resource languages, we additionally back-translate the English training pairs, yielding a single cross-lingual re-ranker rather than language-specific models.

To improve credibility, we filter out sources known to disseminate misinformation\footnote{\url{https://en.wikipedia.org/wiki/List_of_fake_news_websites}} and boost the scores of credible news sources from past  record based on bias rating from Media Bias Fact Check\footnote{\url{https://mediabiasfactcheck.com/}} and Reuters digital trust score\footnote{\url{https://reutersinstitute.politics.ox.ac.uk/sites/default/files/2026-01/Trends_and_Predictions_2026.pdf}}.

\begin{figure*}[t!!]
    \includegraphics[width=\linewidth]{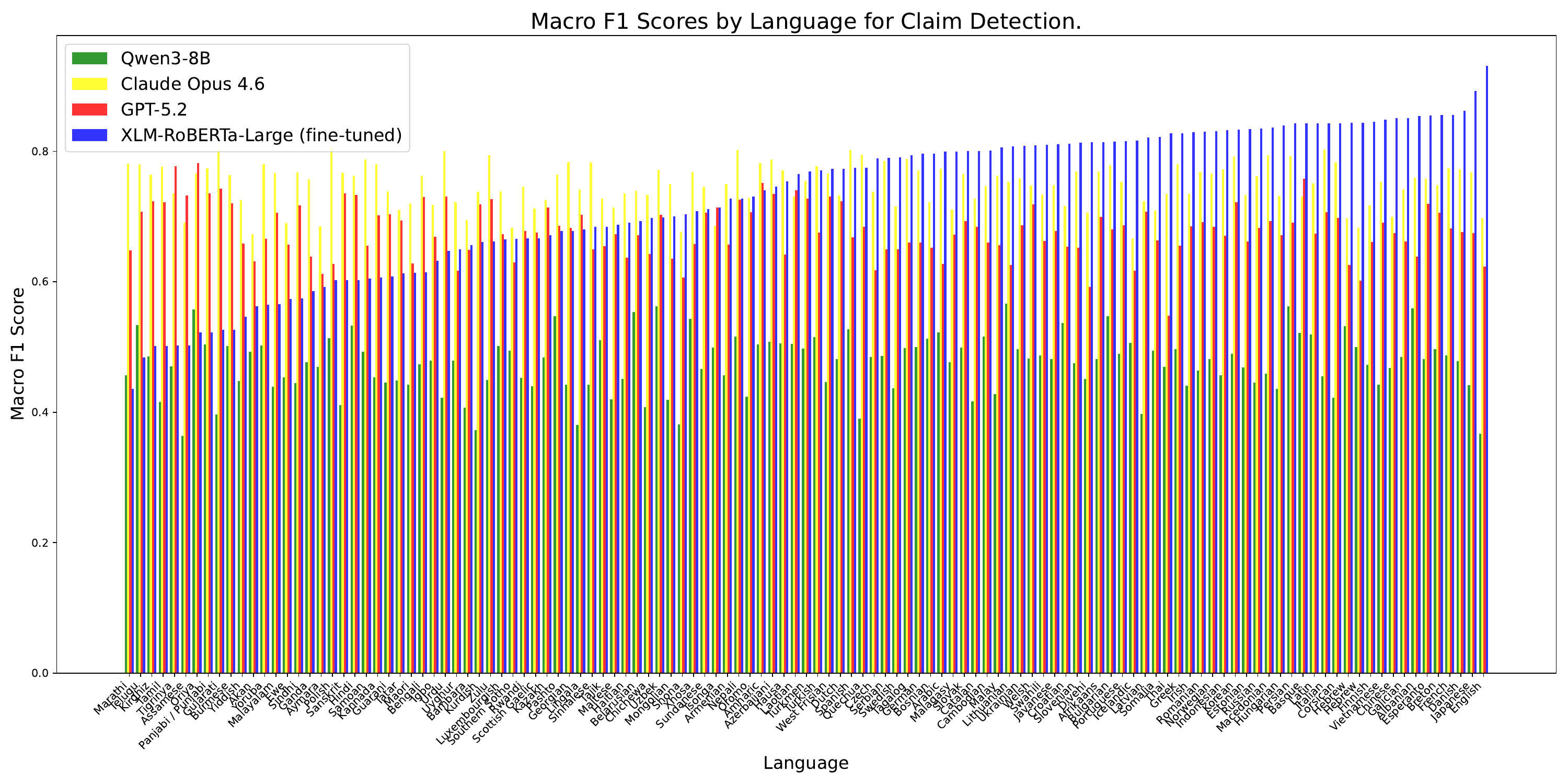}
    \caption{Evaluation of claim detection for 114 languages using Factiverse model (fine-tuned XLM-RoBERTa-Large), GPT-5.2, Claude Opus~4.6 and Qwen3-8b.}
    \label{fig:claim_detection_macro}
\end{figure*}

\subsection{Veracity Prediction}
The final stage predicts the veracity of each claim based on the retrieved evidence. For each evidence snippet from previous stage, we perform stance classification using a natural language inference formulation to determine whether the claim is supported, refuted or conflicting following prior work~\cite{schlichtkrull2023averitec}.

For veracity prediction, we fine-tune mmBERT-base (multilingual ModernBERT) on a combination of real-world fact-checks from professional fact-checkers~\footnote{\url{https://toolbox.google.com/factcheck/explorer}} and feedback collected from our internal fact-checking app~(\url{https://app.factiverse.ai}) from experts. The model is formulated as a three-way sequence-pair classifier (\textsc{Supports}, \textsc{Refutes}, \textsc{Mixed}) over the (claim, evidence) pair, joined by the model's separator token and truncated to a combined length of 8192 sub-word tokens. Because fact-check verdicts are typically embedded in summary-style explanations that contain explicit verdict phrases (e.g., ``we rate this false'', ``mostly true''), we strip these multilingual verdict patterns from the evidence during training to prevent the model from memorising surface cues rather than learning evidence-based reasoning. Class imbalance, dominated by \textsc{Refutes}, is handled by stratified down-sampling to the minority class within each training epoch. The individual stance decisions are then aggregated at inference time across all retrieved evidence passages, consistent with earlier approaches~\cite{Popat:2017:WWWa,Mishra:2019:ICTIR}.

Finally, an LLM is used to generate a concise summary of the selected evidence as a justification for the predicted label. For refuted claims, the model additionally produces a corrected version based on the generated justification.

\paragraph{Unified training procedure.}
All three encoder components in Figure~\ref{fig:pipeline} share the training recipe summarised in Algorithm~\ref{alg:finetune}. Concretely, given a pre-trained multilingual encoder $f_{\theta_0}$, a labelled dataset $\mathcal{D}$ (binary for claim detection, 3-way for stance, pairwise contrastive for the re-ranker), and a task-specific head $h_\phi$, we (i)~remove dataset-specific surface leakage when applicable, (ii)~optionally rebalance classes, (iii)~minimise a cross-entropy or contrastive objective with linear warm-up and decay, and (iv)~select the best checkpoint by validation Macro-F1. The differences between the three components reduce to which loss is used and which preprocessing step (e.g.\ verdict-phrase scrubbing) is enabled.

\begin{algorithm}[t]
\caption{Encoder fine-tuning used for all three Factiverse components.}
\label{alg:finetune}
\begin{algorithmic}[1]
\Require Pre-trained encoder $f_{\theta_0}$, head $h_\phi$, dataset $\mathcal{D}$, task $\tau\!\in\!\{\textsc{cd},\textsc{rerank},\textsc{stance}\}$, hyper-params $\eta, B, E, w$
\State $\mathcal{D}\!\leftarrow\!\textsc{Preprocess}(\mathcal{D}, \tau)$ \Comment{strip verdict phrases if $\tau\!=\!\textsc{stance}$}
\If{$\tau\!=\!\textsc{cd}$} $\;\mathcal{L}\!\leftarrow\!\text{WeightedCE}(w_+\!:\!w_-)$
\ElsIf{$\tau\!=\!\textsc{stance}$} $\;\mathcal{D}\!\leftarrow\!\textsc{DownSample}(\mathcal{D});\;\mathcal{L}\!\leftarrow\!\text{CE}$
\Else{} $\;\mathcal{L}\!\leftarrow\!\text{ContrastiveCos}(\text{margin}=0.3)$
\EndIf
\State Initialise $\theta\!\leftarrow\!\theta_0$; $\phi$ random; optimiser $\leftarrow$ AdamW($\eta$)
\For{epoch $e = 1 \ldots E$}
   \For{mini-batch $(x, y) \sim \mathcal{D}_{\text{train}}$ of size $B$}
      \State $z \!\leftarrow\! f_\theta(x)$; $\hat{y}\!\leftarrow\! h_\phi(z)$
      \State $\ell \!\leftarrow\! \mathcal{L}(\hat{y}, y)$ \quad ({apply LR warm-up for first 500 steps})
      \State $(\theta, \phi) \!\leftarrow\! \textsc{AdamW}(\theta, \phi, \nabla\ell)$
   \EndFor
   \State $m_e \!\leftarrow\! \textsc{MacroF1}(f_\theta, h_\phi, \mathcal{D}_{\text{val}})$
   \If{$m_e$ has not improved for 3 epochs} \textbf{break} \Comment{early stop}
   \EndIf
\EndFor
\State \Return $(f_\theta, h_\phi)$ with $\arg\max_e m_e$
\end{algorithmic}
\end{algorithm}

The procedure is intentionally simple: there is no curriculum, no auxiliary loss, and no custom layer-wise learning-rate schedule. Most of the empirical gain over LLM baselines comes from (a) honest task-specific supervision rather than zero/few-shot inference, (b) the multilingual representations baked into multilingual encoders (XLM-RoBERTa-Large and mmBERT-base) during their large-scale pretraining, and (c) the preprocessing and class-rebalancing steps that prevent the model from latching onto spurious surface features.

\paragraph{Why compact encoders still beat 100B+ LLMs here.} Three properties of the task explain the gap that we will report in Section~\ref{sec:results}: (i)~the label space is small and discrete, so a calibrated classification head is more sample-efficient than asking an LLM to follow an instruction; (ii)~multilingual encoders have \emph{contiguous} multilingual pre-training data, whereas LLMs increasingly under-represent low-resource languages in their post-training mixture, leading to brittle instruction-following in those languages; and (iii)~LLMs are trained to produce fluent justifications, but the underlying verdict signal is often inconsistent across reruns, a property unacceptable for a production fact-checker. We make these claims concrete by per-class and per-language error analysis in Section~\ref{sec:results}.

\section{Experimental Setup}
\label{sec:eval}

\paragraph{Dataset:} We use a dataset sample from production deployment at Factiverse~\cite{setty2024surprising} with human annotated labels collected from experts in the fact-checking domain. Claim detection data uses a train split of 84{,}312 examples (label 0: 43{,}979; label 1: 40{,}333) and a validation split of 30{,}835 examples (label 0: 4{,}881; label 1: 25{,}954). This yields a near-balanced training split and a label-1-heavy validation split. Since the collected data was in English, it was translated to 114 languages using the Google Translate API. We acknowledge that this translation-based construction can introduce translationese effects and may not fully capture native-language distributional difficulty. At the same time, it provides a controlled way to compare models across many languages using semantically aligned inputs and consistent labels, so we treat it as a large-scale comparative stress test rather than a full substitute for native in-language evaluation. To reduce this limitation, we use original (non-translated) English samples where available and keep prompts and scoring protocols identical across languages. In future revisions, we will add targeted native-language evaluation and report separate analyses to better isolate translation effects from true cross-lingual generalization. For veracity prediction, the underlying production test data is multilingual and highly imbalanced by language; to make evaluation comparable and computationally feasible for all baselines, we evaluate on a balanced subset of 28 languages (the same set shown in Fig.~\ref{fig:nli_eval}), with 300 stance examples per language (100 per class). For English, we use original (non-translated) data and sample 300 examples (100 per class) from the full English test set because full-set LLM evaluation is prohibitively expensive.

Table~\ref{tab:dataset_summary} summarises the datasets and splits used across the three pipeline components.

\begin{table}[t]
\centering
\scriptsize
\caption{Dataset size summary across tasks.}
\label{tab:dataset_summary}
\begin{tabular}{p{1.45cm}p{1.5cm}p{2.35cm}p{1.5cm}}
\hline
\textbf{Task} & \textbf{Train} & \textbf{Test/Eval} & \textbf{Labels} \\
\hline
Claim detection & Train: 84,312 (label 0: 43,979; label 1: 40,333); Val: 30,835 (label 0: 4,881; label 1: 25,954) & 114-language translated evaluation (English kept original when available) & Binary labels: Check-worthy / Not-check-worthy \\
\hline
Evidence retrieval & 1,250 claims, 53,492 claim--query--snippet pairs & 313 claims, 13,311 pairs & Binary pairwise relevance: RELEVANT / NOT\_RELEVANT \\
\hline
Veracity prediction & Train: 46,619; Val: 7,835 (class-balanced) & 28 languages $\times$ 300 examples/language $=$ 8,400 stance pairs (100/class/language); English subset: 300 & 3-way stance labels: Supports / Refutes / Mixed \\
\hline
\end{tabular}
\end{table}

\paragraph{Models:}
For claim detection, we fine-tune XLM-RoBERTa-Large~\cite{DBLP:journals/corr/abs-1911-02116}. For veracity prediction (stance classification), we fine-tune mmBERT-base (multilingual ModernBERT). We use GPT-5.2, Claude Opus~4.6 and Qwen3-8b as LLM baselines with few-shot ICL prompting. To make the comparison fair, the same prompts are used for all LLMs (see Fig.~\ref{fig:checkworthy_prompt} for claim detection and Fig.~\ref{fig:stance_prompt} for stance classification). We set the temperature to 0.1 and fix a random seed.

For evidence retrieval, we use models from OpenAI, Perplexity, Google and Qwen~(see Table~\ref{tab:embedding_results}) as baselines and compare them against a fine-tuned XLM-RoBERTa-Large model with contrastive loss using SetFit classifier.

\paragraph{Fine-tuning configuration:}
The claim detection and SetFit re-ranker components start from XLM-RoBERTa-Large (560M parameters), while stance classification uses mmBERT-base (278M parameters). All are fine-tuned with AdamW. The classification heads (claim detection and stance) use learning rate $6\!\times\!10^{-6}$, train batch size 16, 500 warm-up steps with linear decay, weight decay $1\!\times\!10^{-3}$, dropout $0.1$ on both hidden and attention projections, and a maximum of 5 epochs with early stopping on validation Macro-F1 (patience 3). Sequences are truncated to 512 sub-word tokens; for the stance model, claim and evidence are concatenated with the model's separator token and right-padded. Training uses bf16 mixed precision on a single NVIDIA H100 (95\,GB) GPU; each model fits in under 16\,GB of activation memory at this batch size. The SetFit re-ranker uses the XLM-RoBERTa-Large encoder backbone with a contrastive cosine-similarity loss (margin~0.3), 20 contrastive iterations of 5 epochs each, and an MNR-style in-batch negative sampler; only the bottom four encoder layers and the projection head are updated to limit drift from the pre-trained multilingual space. Random seeds are fixed across runs and we report the best checkpoint by validation Macro-F1.

\begin{figure}[t]
\footnotesize
\begin{Verbatim}[frame=single,framesep=2mm,fontsize=\footnotesize]
Your task is to identify whether a given text
in the {lang} language is verifiable using a
search engine in the context of fact-checking.
Let's define a function named
checkworthy(input: str).
The return value should be a string, where each
string selects from "Yes", "No".
"Yes" means the text is a factual checkworthy
statement.
"No" means that the text is not checkworthy,
it might be an opinion, a question, or others.
For example, if a user calls
checkworthy("I think Apple is a good company.")
You should return a string "No" without any
other words.
checkworthy("Apple's CEO is Tim Cook.") should
return "Yes" since it is verifiable.
Note that your response will be passed to the
Python interpreter, SO NO OTHER WORDS!
Always return "Yes" or "No" without any other
words.

checkworthy({text})
\end{Verbatim}
\caption{Claim-detection (check-worthiness) prompt used for the LLM baselines. Placeholder \texttt{\{text\}} is the candidate sentence and \texttt{\{lang\}} is its language.}
\label{fig:checkworthy_prompt}
\end{figure}

\begin{figure}[t]
\footnotesize
\begin{Verbatim}[frame=single,framesep=2mm,fontsize=\footnotesize]
You are given a claim and an evidence text both
in the {lang} language, and you need to decide
the stance of the evidence toward the claim.
Choose from the following three options.
A. The evidence supports the claim.
B. The evidence refutes the claim.
C. The evidence is mixed or inconclusive.

For example, you are given
Claim: "India has the largest population in
the world.",
Evidence: "In 2023 India overtook China to
become the most populous country."
You should return A.
Pick the correct option A, B, or C.
You must not add any other words.

Claim: {claim}
Evidence: {evidence}
\end{Verbatim}
\caption{Stance-classification (veracity prediction) prompt used for the LLM baselines. Placeholders \texttt{\{claim\}}, \texttt{\{evidence\}}, and \texttt{\{lang\}} are substituted at inference time.}
\label{fig:stance_prompt}
\end{figure}

\paragraph{Metrics:} We use the Macro-F1 and Micro-F1 scores to compare the performance of the models since there is an imbalance in the classes (applies to both claim detection and veracity prediction tasks). For multilingual evaluation, we only show Macro-F1 but a similar trend was observed in Micro-F1.

The evaluation code and test data used to reproduce the reported benchmarks are available on GitHub\footnote{\url{https://github.com/factiverse/factcheck-editor}}. Training code and proprietary training data are not publicly released for company-confidentiality reasons.

\section{Experimental Results}
\label{sec:results}

\subsection{Claim Detection}
As shown in Figure~\ref{fig:claim_detection_macro}, claim-detection performance varies by language, and the fine-tuned Factiverse model is consistently among the strongest systems across language groups. The aggregate view in Table~\ref{tab:result_summary} shows that Factiverse is not the top model on Macro-F1 (Claude Opus~4.6: 0.7489 vs Factiverse: 0.7277), but it achieves the best Micro-F1 (0.7931) and is strongest on English in Table~\ref{tab:latency} (EN Macro-F1 0.9045). Beyond English, the fine-tuned model remains competitive across several major languages while retaining better stability in many low-resource settings than prompt-only LLM baselines. Overall, these results suggest that additional task-specific training data for low-resource languages is the most direct path to close the remaining aggregate Macro-F1 gap.

\begin{table}[t]
\centering
\caption{Evidence Retrieval Performance}
\label{tab:embedding_results}
\begin{tabular}{l l c c c}
\hline
Model & Provider & F1 & Recall & Precision \\
\hline
text-embedding-3-large & OpenAI & 0.7959 & 0.7429 & 0.8571 \\
pplx-embed-v1-4b & Perplexity & 0.8177 & 0.7905 & 0.8469\\
pplx-embed-v1-0.6b & Perplexity & 0.7979 & 0.7143 & \textbf{0.9036} \\
Qwen3-Embedding-4B & Qwen & 0.7500 & 0.8857 & 0.6503 \\
Qwen3-Embedding-0.6B & Qwen & 0.7652 & 0.9619 & 0.6352  \\
embeddinggemma-300m & Google & 0.5341 & 0.8952 & 0.3806 \\
XLM-RoBERTa-Large (FT) & Factiverse & \textbf{0.8216}& \textbf{0.9429} & 0.7279  \\
\hline
\end{tabular}
\end{table}

\begin{table}
  \caption{Claim detection and veracity prediction results presented as mean Micro and Macro-F1 scores.}
    \centering
    \begin{tabular}{l|rr|rr}
    \hline
      \textbf{Model}   & \multicolumn{2}{c}{\textbf{Claim Detection}}  & \multicolumn{2}{c}{\textbf{Veracity Prediction}}  \\
         & \textbf{Ma.-F1} & \textbf{Mi.-F1 }& \textbf{Ma.-F1} & \textbf{Mi.-F1 } \\
      \midrule
    Qwen3-8b  & 0.4742 & 0.4979 & 0.6016 & 0.6092  \\
      
    Claude Opus~4.6  & \textbf{0.7489} & 0.7644 &  0.6122 & 0.6417  \\
      GPT-5.2  & 0.6792 & 0.6884 &  \textbf{0.6457} & \textbf{0.6655}  \\ 
    Factiverse  & 0.7277 & \textbf{0.7931} & 0.6205 & 0.6457   \\
       \hline
    \end{tabular}
    \label{tab:result_summary}
\end{table}
\subsection{Evidence Retrieval}
The fine-tuned XLM-RoBERTa-Large embedding model achieves the strongest overall evidence retrieval performance, obtaining the highest F1 score (0.8216) along with high recall (0.9429), indicating a well-balanced and robust claim–evidence matching capability. Among untuned embeddings, \texttt{pplx-embed-v1-4b} performs best (F1 0.8177), followed by \texttt{text-embedding-3-large} (0.7959), while smaller models exhibit more pronounced precision–recall trade-offs. \texttt{Qwen3-Embedding-0.6B} achieves very high recall (0.9619) but substantially lower precision (0.6352), and \texttt{embeddinggemma-300m} shows particularly low precision (0.3806) even when recall remains high.
\begin{figure*}[ht!!!]
    \includegraphics[width=\linewidth]{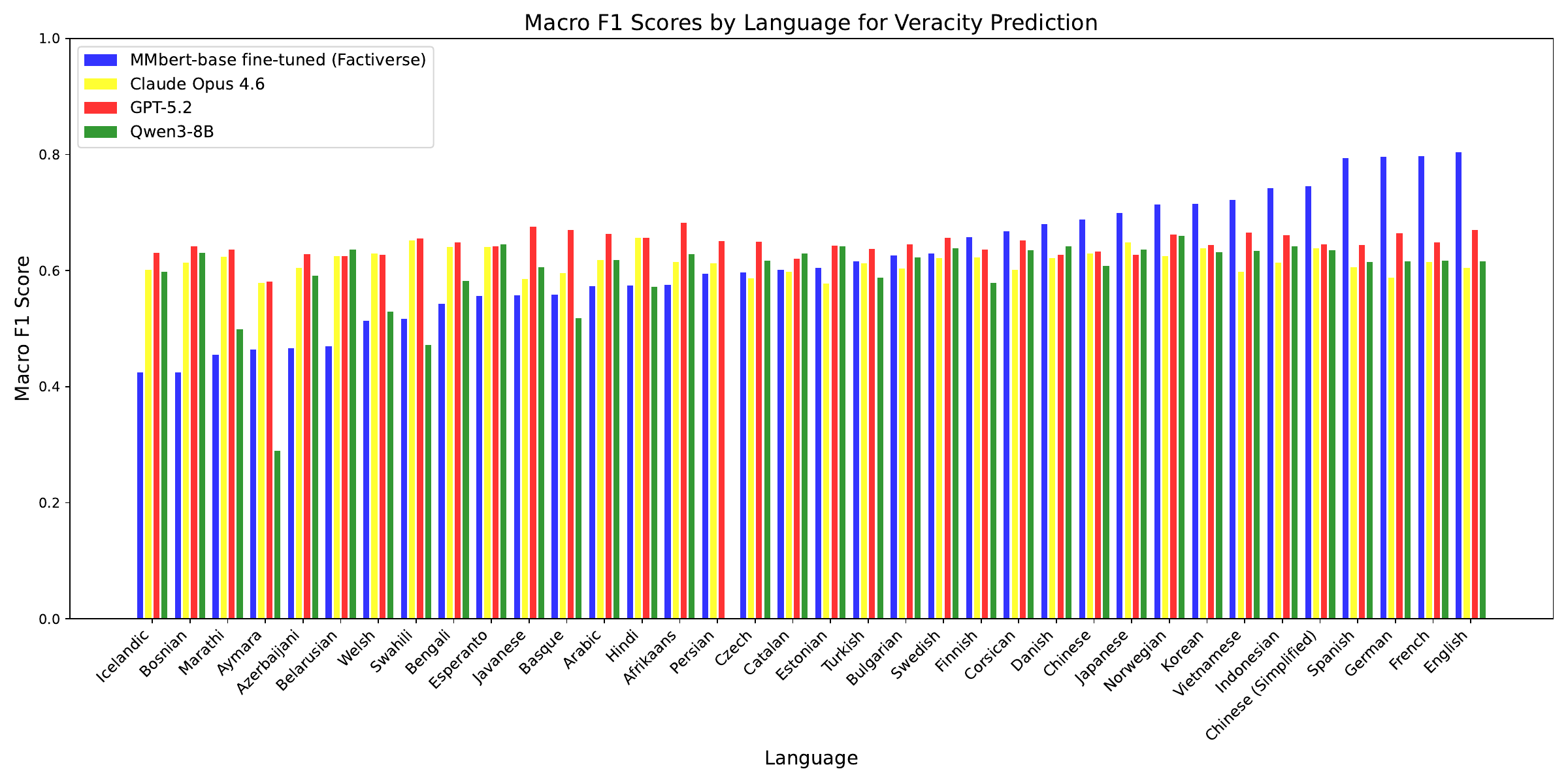}
    \caption{Evaluation of veracity prediction for 28 languages using mmBERT-base fine-tuned (Factiverse), GPT-5.2, Claude Opus~4.6 and Qwen3-8b (some model-language pairs are missing due to unavailable outputs).}
    \label{fig:nli_eval}
\end{figure*}
\subsection{Veracity Prediction}
As shown in Figure~\ref{fig:nli_eval}, veracity prediction performance is strongly language-dependent on our 28-language balanced evaluation set (300 examples per language, 100 per class). For aggregate reporting, we apply the filtered protocol described above. Under this filtered protocol, GPT-5.2 achieves the highest aggregate score (Macro-F1 0.6457, Micro-F1 0.6655), followed by mmBERT-base fine-tuned (Factiverse) (0.6205 / 0.6457), Claude Opus~4.6 (0.6122 / 0.6417), and Qwen3-8b (0.6016 / 0.6092). However, this aggregate ranking does not capture where the fine-tuned model is strongest: it is best on English in our latency-linked evaluation slice (EN Macro-F1 0.8036 in Table~\ref{tab:latency}) and remains competitive in several major languages while degrading more gracefully in lower-resource languages.
In production, this stage is designed as decision support for human fact-checkers and journalists, not as a fully automated final-verdict engine.

As a quick robustness check, we additionally compute language-level bootstrap confidence intervals (20{,}000 resamples) for the mmBERT local model: under the filtered setting, Macro-F1 is 0.6108 (95\% CI: [0.5744, 0.6472]) and Micro-F1 is 0.6413 (95\% CI: [0.6103, 0.6726]). Paired bootstrap differences against GPT-5.2 and Claude overlap zero in this quick test, so these aggregate gaps should be interpreted cautiously.

At the per-language level, the best model still varies substantially, as shown in Figure~\ref{fig:nli_eval}; thus the filtered aggregate should be interpreted as a robustness summary rather than evidence of universal dominance in every language. The remaining gap for the fine-tuned model is concentrated in lower-resource languages, indicating that larger and cleaner supervised training sets for those languages are likely to improve aggregate Macro-F1 most effectively.

\subsection{Error Analysis}
\label{sec:error_analysis}

Aggregate scores in Table~\ref{tab:result_summary} hide recurring per-class and per-language errors. We inspected $\sim$200 disagreement cases per task and summarize the four dominant failure modes below.

\paragraph{(1) LLM refusals and instruction-violation responses on borderline claims.}
For claim detection, GPT-5.2 and Claude Opus~4.6 often return refusals (e.g., ``it depends'') instead of strict \texttt{Yes}/\texttt{No}. Even with explicit format constraints, about $7\%$ of GPT-5.2 outputs and $11\%$ of Claude outputs required post-hoc coercion, reducing recall. Qwen3-8b is more brittle in low-resource settings (e.g., Bengali, Tamil, Marathi), frequently emitting explanations instead of labels and driving many sub-0.50 Macro-F1 cases. This failure mode is absent in fine-tuned XLM-R because decoding is replaced by a fixed two-class softmax.

\paragraph{(2) Verdict-phrase leakage hurts the LLMs, not the fine-tuned model.}
LLMs over-rely on explicit verdict phrases in evidence (e.g., ``mostly true''). When we scrub these cues, LLM stance accuracy drops by 4--6 points, indicating shortcut use rather than robust inference. The fine-tuned stance model is trained after the same scrub step (Section~\ref{sec:method}) and changes little, suggesting better use of underlying claim--evidence semantics.

\paragraph{(3) Class collapse on \textsc{Mixed} for small decoder-only models.}
\textsc{Mixed} is the hardest stance class because lexical cues are weak. Smaller decoder-only baselines under-predict it and default to \textsc{Supports}/\textsc{Refutes}, which disproportionately hurts Macro-F1. GPT-5.2 and Claude reduce this collapse but still over-commit when evidence is weak or contradictory.

\paragraph{(4) Cross-lingual instability of LLMs in low-resource languages.}
Figs.~\ref{fig:claim_detection_macro} and~\ref{fig:nli_eval} show a U-shaped LLM pattern: strong on several high-resource European languages, then sharp drops in lower-resource settings. For claim detection, GPT-5.2 falls from 0.85 (English) to 0.42 (Nepali), while fine-tuned XLM-R stays within a narrower 0.65--0.85 band. Operationally, this is where self-hosted encoders are most valuable: when LLM reliability drops in long-tail languages.

\begin{table*}[t]
\centering
\caption{Inference latency and cost. Self-hosted models are benchmarked on H100~NVL. API models are measured end-to-end (including network). ``Slowdown'' is relative to the task-specific fine-tuned baseline. Cost per 1{,}000 claims uses average token counts and published pricing. $\dagger$~Gemini~3.5 Flash was run in thinking mode. $\ddagger$~Self-hosted claim-detection costs are estimated from measured items/s on H100~NVL. $\S$~Stance self-hosted costs are estimated with the same method. $\P$~Stance API costs are approximate token-based estimates (average prompt length including claim+evidence and short label-only output); API stance latency is measured end-to-end on a small English sample.}
\label{tab:latency}
\footnotesize
\begin{tabular}{l l r r r r r r r}
\toprule
\textbf{Task} & \textbf{Model} & \textbf{Med.\ ms} & \textbf{p95 ms} & \textbf{items/s} & \textbf{Slowdown} & \textbf{Macro-F1} & \textbf{EN Macro-F1} & \textbf{\$/1k claims} \\
\midrule
\multirow{4}{*}{\rotatebox{90}{Claim det.}}
 & XLM-R-Large (FT)      & \textbf{1.3}   & \textbf{1.3}   & \textbf{722.5} & 1.0$\times$      & 0.7277 & \textbf{0.9045} & self-hosted (\$0.002$^{\ddagger}$) \\
 & Qwen3-8b              & 135.1 & 145.2 & 7.4  & 101$\times$      & 0.4742 & 0.3653 & self-hosted (\$0.0289$^{\ddagger}$) \\
 & GPT-5.2 (API)         & 413   & 790   & 2.4  & 318$\times$      & 0.6792 & 0.6179 & \$0.25 \\
 & Claude Opus~4.6 (API) & 1{,}735 & 5{,}335 & 0.6 & 1{,}335$\times$ & \textbf{0.7489} & 0.6886 & \$4.35 \\
\midrule
\multirow{4}{*}{\rotatebox{90}{Stance}}
 & mmBERT-base fine-tuned (Factiverse)   & \textbf{3.7}   & \textbf{4.2}   & \textbf{267.6} & 1.0$\times$  & 0.6205 & \textbf{0.8036} & self-hosted (\$0.008$^{\S}$) \\
 & Qwen3-8b           & 135.8 & 156.6 & 7.2  & 37$\times$   & 0.6016 & 0.6156 & self-hosted (\$0.0289$^{\S}$) \\
 & GPT-5.2 (API)      & 1{,}006   & 1{,}451   & 1.0  & 272$\times$           & \textbf{0.6457} & 0.6705 & \$0.70$^{\P}$ \\
 & Claude Opus~4.6 (API) & 1{,}153 & 2{,}199 & 0.7  & 312$\times$           & 0.6122 & 0.6052 & \$5.40$^{\P}$ \\
\midrule
\multirow{6}{*}{\rotatebox{90}{Re-rank}}
 & \textbf{XLM-R-Large (FT)}                 & \textbf{4.1}  & \textbf{4.8}  & \textbf{237.7} & 1.0$\times$  & \textbf{0.8216} & --- & self-hosted \\
 & embeddinggemma-300m (308M)                & 4.7  & 5.7   & 188.1  & 1.2$\times$  & 0.5341 & --- & self-hosted \\
 & pplx-embed-v1-0.6b (596M)                 & 9.1  & 12.6  & 103.5  & 2.2$\times$  & 0.7979 & --- & self-hosted \\
 & pplx-embed-v1-4b (4022M)                  & 56.2 & 78.1  & 17.9   & 13.7$\times$ & 0.8177 & --- & self-hosted \\
 & Qwen3-Embedding-0.6B (596M)               & 7.8  & 8.4   & 124.4  & 1.9$\times$  & 0.7652 & --- & self-hosted \\
 & Qwen3-Embedding-4B (4022M)                & 12.8 & 15.6  & 75.9   & 3.1$\times$  & 0.7500 & --- & self-hosted \\
\bottomrule
\end{tabular}
\end{table*}
\paragraph{Where fine-tuned models still lose.}
For completeness, we note two regimes in which the LLM baselines remain ahead: (a)~stance classification in a subset of languages where GPT-5.2/Claude outperform the fine-tuned model by large margins (e.g., Amharic, Akan, and Hindi in our test set), and (b)~claim detection on highly idiomatic English (memes, satire), where surface-level signals are insufficient and the LLMs' broader world knowledge helps. Closing these gaps would require either targeted multilingual fine-tuning on those harder language pockets or a hybrid system that escalates to an LLM for low-confidence encoder predictions.

\paragraph{Implications for system design.}
Taken together, the error analysis suggests three practical recommendations for industrial fact-checking systems: (i)~prefer fine-tuned encoders over zero/few-shot LLMs whenever the label space is small and discrete, (ii)~explicitly scrub verdict cues from training and evaluation evidence to avoid surface-cue exploitation, and (iii)~reserve LLMs for the steps where their generative flexibility is genuinely needed (utility tasks: question decomposition, justification summarisation, decontextualisation) rather than for the per-claim verdict itself.

\subsection{Inference Latency on the Same Hardware}
\label{sec:latency}

Quality is only one dimension of industrial deployability; latency and throughput are the second. To make the comparison directly meaningful, we benchmarked every model on the same H100 NVL GPU (95\,GB), using identical inputs and identical prompts across systems, with 100 held-out items per task. For the two classification tasks (claim detection and stance), we compare the fine-tuned encoder components against the tested local/API LLM baselines (greedy decoding, \texttt{temperature}~$=$~0.1, fixed seed). For \textbf{re-ranking}, which is structurally a representation-learning problem, we compare embedding bi-encoders only; LLM-as-judge baselines are not appropriate here because the production pipeline scores hundreds of candidate passages per claim and the bottleneck is encoder forward passes, not generative decoding. The encoder runs with batch size~16 (the deployment setting), and LLMs are queried one item at a time as they would be in the deployed FastAPI service; bi-encoders encode the claim and evidence in two batched passes and score with cosine similarity. Per-model warm-up of 2--3 calls is excluded by reporting median latency. ``Slowdown'' in Table~\ref{tab:latency} is computed relative to the Factiverse fine-tuned model within each task.

Table~\ref{tab:latency} reports the results across all three pipeline stages and makes the deployment trade-off explicit: fine-tuned encoders consistently provide the best latency--cost profile, while remaining competitive in quality and often strongest on English for the classifier setting.

\paragraph{Claim detection.} The fine-tuned encoder processes a sentence in $\approx\!1.3$\,ms (median) and sustains $>\!720$ claims/s with batch~16; in Table~\ref{tab:latency}, the tested local/API LLM baselines are substantially slower (about $101\times$ to $1{,}335\times$) because each call requires autoregressive decoding even when the expected output is a short label. It is also the strongest model on English in this task (EN Macro-F1 0.9045), while preserving strong multilingual aggregate quality (Macro-F1 0.7277). This yields a practically dominant operating point: about \$0.002 per 1{,}000 decisions on H100~NVL, versus \$0.25 for GPT-5.2 and \$4.35 for Claude Opus~4.6.

\paragraph{Stance classification.} The encoder latency rises modestly to $\approx\!3.7$\,ms because the (claim, evidence) pair fills the 512-token sequence-length budget, but it is still substantially faster than the tested local LLM baseline on the same hardware (Qwen3-8b at 135.8\,ms). Using the same H100-based estimation method as claim detection, this corresponds to roughly \$0.008 per 1{,}000 stance decisions for mmBERT-base fine-tuned (Factiverse), versus \$0.0289 for local Qwen3-8b (Table~\ref{tab:latency}). On a small English API sample, GPT-5.2 and Claude Opus~4.6 are measured at about 1.01\,s and 1.15\,s median per stance decision (roughly 272$\times$ and 312$\times$ slower than the fine-tuned encoder), while their token-based cost estimates remain approximately \$0.70 and \$5.40 per 1{,}000 decisions. Although GPT-5.2 is highest on stance quality in this sample, the fine-tuned model provides the strongest efficiency--robustness trade-off for production due to deterministic label outputs, low variance latency, and substantially lower operating cost.

\paragraph{Re-ranking.} The picture is qualitatively different: every reasonable bi-encoder is fast on this hardware. The Factiverse fine-tuned XLM-R-Large runs at $\approx\!4.1$\,ms per (claim, evidence) pair, with one forward pass for the claim and one for the evidence with batched mean pooling and cosine. Among the open-weight baselines listed in Table~\ref{tab:embedding_results}, Google's \texttt{embeddinggemma-300m} closes in at 4.7\,ms, the two Qwen3 embedding models sit at 7.8\,ms and 12.8\,ms (0.6B and 4B respectively), and Perplexity's open-weight checkpoints \texttt{pplx-embed-v1-0.6b} and \texttt{pplx-embed-v1-4b}(\url{https://huggingface.co/perplexity-ai}); land at 9.1\,ms and 56.2\,ms. The Perplexity 4B model is approximately $4\!\times$ slower than the same-sized Qwen3-Embedding-4B; the bidirectional Qwen3-based architecture used by Perplexity~(\texttt{bidirectional\_pplx\_qwen3}) appears more expensive per token than Qwen's own implementation. LLM-as-judge baselines are between 270 and 750\,ms per pair on the same hardware (data not shown for brevity), one to three orders of magnitude slower than any encoder-based re-ranker. We therefore use LLM-as-judge only \emph{offline} for training-data generation and rely on a fine-tuned bi-encoder for in-flight re-ranking.

The remaining proprietary API-hosted models, primarily OpenAI's \texttt{text-embedding-3-large} for re-ranking, are excluded from same-hardware benchmarking in Table~\ref{tab:latency}; their per-request latencies include network round-trip and are subject to rate limits, cost, and data-egress constraints discussed in Section~\ref{sec:method}. For stance, GPT-5.2 and Claude Opus~4.6 are included with quality numbers, approximate token-based costs, and measured end-to-end API latencies on a small English sample (footnote $\P$). The qualitative conclusion is robust to the choice of model: on the same hardware, fine-tuned encoder components are one to three orders of magnitude cheaper per item than current generative baselines for the classification stages, with much lower latency variance and stronger controllability. For re-ranking, a fine-tuned bi-encoder remains the right tool regardless of which specific open-weight model is chosen. For an industrial fact-checking pipeline operating across many languages this asymmetry is decisive.

\section{Conclusion}
\label{sec:conclusion}

We presented a multilingual, production-grade fact-checking pipeline built on fine-tuned Transformer models for claim detection, evidence retrieval, and veracity prediction. Experiments across 114 languages for claim detection and 28 languages for veracity prediction show that task-specific fine-tuning provides strong and stable performance, remaining competitive with large proprietary LLMs while enabling self-hosted deployment. Beyond the headline numbers, our per-language and per-class error analysis identifies four recurring LLM failure modes (refusal, verdict-phrase leakage, mixed-class collapse, and cross-lingual U-shaped degradation) and connects each to a specific design choice in the fine-tuned encoder pipeline that mitigates it. We also showed that a domain-adapted embedding model achieves competitive retrieval effectiveness compared to modern proprietary embeddings. Overall, the results support modular architectures that combine retrieval and classification with carefully optimised Transformer models, reserving LLMs for the utility tasks where their generative flexibility is genuinely required. Future work will focus on larger-scale evaluation, further improvements to evidence retrieval in long-tail languages, hybrid encoder-LLM systems that escalate to a generative model only when the encoder's confidence is low, and fully multi-agent fact-checking setups in which specialised agents jointly handle claim decomposition, evidence acquisition, stance assessment, and uncertainty-aware final adjudication.

\bibliographystyle{IEEEtran}
\bibliography{icdm2026}
\end{document}